\documentclass[runningheads]{llncs}
\usepackage{graphicx}
\usepackage{amsmath,amssymb} 
\usepackage{color}
\usepackage[width=122mm,left=12mm,paperwidth=146mm,height=193mm,top=12mm,paperheight=217mm]{geometry}
\usepackage{array}
\usepackage{multirow}
\usepackage{subfig}
\usepackage{epstopdf}
\usepackage{url}
\usepackage{lipsum}

\newcommand\blfootnote[1]{%
	\begingroup
	\renewcommand\thefootnote{}\footnote{#1}%
	\addtocounter{footnote}{-1}%
	\endgroup
}

\begin{document}
	\pagestyle{headings}
	\mainmatter
	
	\title{Reference-Conditioned Super-Resolution by Neural Texture Transfer} 
	
	\titlerunning{Reference-Conditioned Super-Resolution by Neural Texture Transfer}
	
	\authorrunning{Zhifei Zhang, Zhaowen Wang, Zhe Lin, and Hairong Qi}
	
	\author{Zhifei Zhang \inst{1} \and Zhaowen Wang \inst{2} \and Zhe Lin \inst{2} \and Hairong Qi \inst{1}}
	
	
	\institute{Department of Electrical Engineering and Computer Science,\\
		University of Tennessee, Knoxville.  \email{\{zzhang61,hqi\}@utk.edu} \and  Adobe Research. \email{\{zhawang,zlin\}@adobe.com}
	}
	
	\newcolumntype{L}[1]{>{\raggedright\let\newline\\\arraybackslash\hspace{0pt}}m{#1}}
	\newcolumntype{C}[1]{>{\centering\let\newline\\\arraybackslash\hspace{0pt}}m{#1}}
	\newcolumntype{R}[1]{>{\raggedleft\let\newline\\\arraybackslash\hspace{0pt}}m{#1}}
	\newcommand{\alg}{SRNTT}
	\newcommand{\dataset}{CUFED5}
	\newcommand{\shrink}{\vspace{-15pt}}
	
	\maketitle
	
	\begin{abstract}
		With the recent advancement in deep learning, we have witnessed a great progress in single image super-resolution. However, due to the significant information loss of the image downscaling process, it has become extremely challenging to further advance the state-of-the-art, especially for large upscaling factors. This paper explores a new research direction in super resolution, called reference-conditioned super-resolution, in which a reference image containing desired high-resolution texture details is provided besides the low-resolution image.  
		We focus on transferring the high-resolution texture from reference images to the super-resolution process without the constraint of content similarity between reference and target images, which is a key difference from previous example-based methods.
		Inspired by recent work on image stylization, we address the problem via neural texture transfer. We design an end-to-end trainable deep model which generates detail enriched results by adaptively fusing the content from the low-resolution image with the texture patterns from the reference image. \blfootnote{
			Code:  \url{http://web.eecs.utk.edu/~zzhang61/project_page/SRNTT/SRNTT.html}}
		We create a benchmark dataset for the general research of reference-based super-resolution, which contains reference images paired with low-resolution inputs with varying degrees of similarity.
		Both objective and subjective evaluations demonstrate the great potential of using reference images as well as the superiority of our results over other state-of-the-art methods.
	\end{abstract}
	
	\section{Introduction}
	The traditional single image super-resolution (SR) problem is defined as recovering a high-resolution (HR) image from its low-resolution (LR) observation~\cite{yang2014single}, which has received substantial attention in the computer vision community. As in other fields of computer vision studies, the introduction of convolutional neural networks (CNNs)~\cite{dong2014learning,wang2015deep,kim2016deeply,lim2017enhanced} has greatly advanced the state-of-the-art performance of SR. However, due to the ill-posed nature of SR problems, most existing methods still suffer from blurry results at large upscaling factors, e.g., 4$\times$, especially when it comes down to the recovery of fine texture details in the original HR image which are lost in the LR counterpart. 
	In recent years, perceptual-related constraints, e.g., perception loss~\cite{johnson2016perceptual} and adversarial loss~\cite{goodfellow2014generative}, have been introduced to the SR problem formulation, leading to major breakthroughs on visual quality under large upscaling factors~\cite{ledig2017photo,sajjadi2017enhancenet}. However, they tend to add fake textures and even artifacts to make the SR image of visually higher-resolution. 
	\begin{figure}[t]
		\centering
		\begin{tabular}{C{2.5cm}C{2.4cm}C{2.5cm}C{2.5cm}C{1.5cm}C{.1cm}}
			\centering Bicubic & 
			\centering SRCNN & 
			\centering SRGAN & 
			\centering \alg & 
			\centering Reference & ~
		\end{tabular}
		\includegraphics[width=.2\columnwidth]{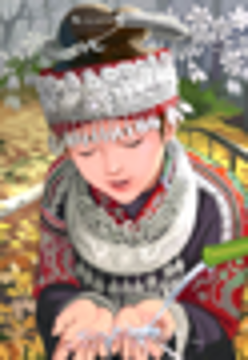}
		\includegraphics[width=.2\columnwidth]{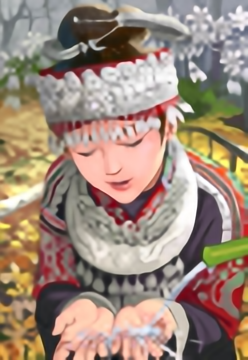}
		\includegraphics[width=.2\columnwidth]{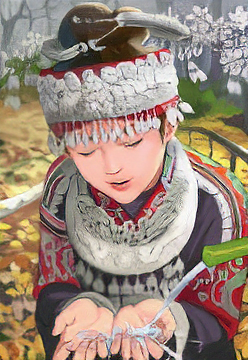}
		\includegraphics[width=.2\columnwidth]{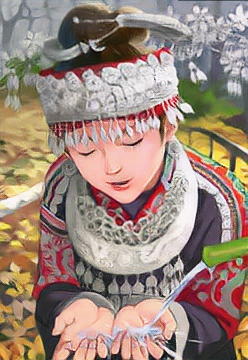}
		\includegraphics[width=.145\columnwidth]{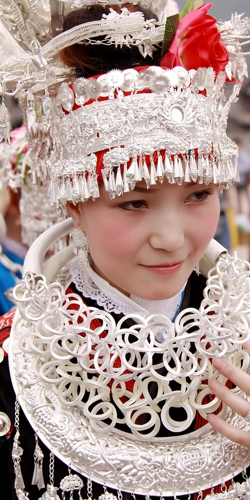}
		\caption{From left to right: bicubic interpolation, SRCNN that minimizes MSR, SRGAN that further incorporates perceptual-related constraints, the proposed \alg~conditioned on the reference shown in the upper-right corner, and the reference image for \alg. The upscaling factor is 4$\times$.}
		\label{fig:demo}
	\end{figure}
	
	This paper diverts from the traditional SR and explores a new research direction --- reference-conditioned super-resolution, which utilizes the rich textures from HR references to compensate for the lost details in the LR images, relaxing the ill-posed issue and producing more detailed and realistic textures with the help of reference images. The reference images may come from photo albums, video frames, or web image search.
	There are existing example-based SR approaches~\cite{freeman2002example,chang2004super,freedman2011image,yue2013landmark,timofte2013anchored,liu2017retrieval} that adopted external high-frequency information to enhance textures. However, they assume the reference images could be well aligned or present similar texture to the LR images. 
	By contrast, the reference image plays a different role in our setting: it does not need to have similar content with target HR image. Instead, we only intend to transfer the relevant texture from reference image to target image. 
	
	Inspired by recent work on image stylization~\cite{gatys2016image,johnson2016perceptual,chen2016fast}, we propose the Super-Resolution by Neural Texture Transfer (SRNTT), which \textit{adaptively} transfers textures to the SR image conditioned on the reference image.
	More specifically, \alg~conducts local texture matching in the high-level feature space and adaptively fuses matched textures with a deep model.
	Fig.~\ref{fig:demo} illustrates the advantage of the proposed \alg~compared with two representative works, SRCNN~\cite{dong2014learning,dong2016image} and SRGAN~\cite{ledig2017photo}, which are without and with perceptual-related constraints, respectively. 
	\alg~shows improved texture transferred from the reference. Note that the texture in reference does not have to match the one in HR ground truth. We emphasize the capacity of SRNTT in handling arbitrary reference images. For example, as shown in Fig.~\ref{fig:weight} with the extreme case of the reference image being simply random noise, the proposed SRNTT is still able to generate the SR image with comparable visual quality as that from SRGAN.
	
	That being said, similarity between the reference and LR image is still the key factor that affects the performance of reference-conditioned SR. However, there is no existing benchmark dataset that could provide different similarity levels of references for the investigation of adaptiveness and robustness. To facilitate fair comparison and further research on the reference-conditioned SR problem, we propose a new dataset, named \dataset, which provides training and testing sets accompanied with references of five similarity levels that vary in content, texture, color, illumination, and view point. For example, as shown in Fig.~\ref{fig:test_set}, the least similar reference for an image of a building could be a person.
	
	The main contributions of this paper are:
	\begin{itemize}
		\item We explore a new research direction of SR, i.e., reference-conditioned super-resolution, as opposed to SISR which only relies on the LR input image, and example-based SR which makes rigid assumptions on the external example image used. 
		Reference-conditioned SR aims to generate HR texture information for LR input image by referring to an arbitrary external image, and thus enables the generation of SR images with plausible texture details even at large upscale factor, further advancing the state-of-the-art in SR research.
		\item We propose an end-to-end deep model, \alg, to recover the LR image conditioned on any given reference. 
		We demonstrate the adaptiveness, effectiveness, and visual improvement of the proposed \alg~by extensive empirical studies.   
		\item We create a benchmark dataset,  \dataset, to facilitate the performance evaluation of SR methods in handling references with different levels of similarity to the LR input image.  
		
	\end{itemize}
	
	In the rest of this paper, we review the related works in Section~\ref{sec:related_work}. The network architecture and training criteria are discussed in Section~\ref{sec:approach}. In Section~\ref{sec:dataset}, the proposed dataset \dataset~is described in detail. The results of quantitative and qualitative evaluations are presented in Section~\ref{sec:experiment}. Finally, Section~\ref{sec:conclusion} concludes this paper.
	
	\section{Related Works}
	\label{sec:related_work}
	
	\subsection{Deep learning based single image SR}
	
	In recent years, deep learning based SISR has shown superior performance in either PSNR or visual quality compared to those non-deep-learning based methods~\cite{dong2014learning,wang2015deep,ledig2017photo}. The reader could refer to~\cite{nasrollahi2014super,yang2014single} for more comprehensive review of SR. Here we only cover deep learning based methods.
	
	A milestone work that introduced CNN into SR was proposed by Dong et al.~\cite{dong2014learning}, where a three-layer fully convolutional network was trained to minimize MSE between the SR image and original HR image. It demonstrated the effectiveness of deep learning in SR and achieved the state-of-the-art performance. Wang et al.~\cite{wang2015deep} combined the strengths of sparse coding and deep network and made considerable improvement over previous models. To speed up the SR process, Dong et al.~\cite{dong2016accelerating} and Shi et al.~\cite{shi2016real} extracted features directly from the LR image, that also achieved better performance compared to processing the upscaled LR image through bicubic. To further reduce the number of parameters, Lai et al.~\cite{lai2017fast} progressively reconstructed the sub-band residuals of HR images at multiple pyramid levels. In recent two years, the state-of-the-art performance (in PSNR) were all achieved by deep learning based models~\cite{kim2016deeply,kim2016accurate,lim2017enhanced}.
	
	The above mentioned methods, in general, aim at minimizing the mean squared error (MSE) between the SR and HR images, which might not always be consistent with the human evaluation results (i.e., perceptual quality)~\cite{ledig2017photo,sajjadi2017enhancenet}. 
	As a result, perceptual-related constraints were incorporated to achieve better visual quality. Johnson et al.~\cite{johnson2016perceptual} demonstrated the effectiveness of adding perception loss using VGG~\cite{simonyan2014very} in SR. Ledig et al.~\cite{ledig2017photo} introduced adversarial loss from the generative adversarial nets (GANs)~\cite{goodfellow2014generative} to minimize the perceptually relevant distance between the SR and HR images. Sajjadi et al.~\cite{sajjadi2017enhancenet} further incorporated the texture matching loss based on the idea of style transfer~\cite{gatys2015texture,gatys2016image}, to enhance the texture in the SR image. The proposed \alg~is more related to \cite{ledig2017photo,sajjadi2017enhancenet}, where perceptual-related constraints (i.e., perceptual loss and adversarial loss) are incorporated to recover more visually plausible SR images.

	\subsection{Example-based SR methods}

	
	In contrast to SISR where only the single LR image is used as input, example-based SR methods introduce additional images to assist the SR process. In general, the example images need to possess very similar texture or content structure with the LR image. The examples could be selected from adjacent frames in a video~\cite{liu2011bayesian,caballero2017real}, images from web retrieval~\cite{yue2013landmark}, or self patches~\cite{freeman2002example,freedman2011image}. We will not discuss video (multi-frame) super-resolution which are specifically designed taking advantage of the similarity nature of adjacent frames. The proposed reference-conditioned SR allows a more relaxed scenario --- the reference could be an arbitrary image.
	
	Those early works~\cite{freeman2002example,chang2004super} not using deep learning mostly built the mapping from LR to HR patches and fused the HR patches at the pixel level and by a shallow model, which is insufficient to model the complicated dependency between the LR image and extracted details from the HR patches, i.e., examples. In addition, they implied that each LR patch could be matched to an appropriate HR patch (similar textures always present in the example). 
	Freedman and Fattal~\cite{freedman2011image} and Huang et al.~\cite{Huang-CVPR-2015} referred to self examples for similar textures and used a shallow model for texture transfer. 
	A more generic scenario of utilizing the examples was proposed by Yue et al.~\cite{yue2013landmark}, which instantly retrieved similar images from web and conducted global registration and local matching. However, they made a strong assumption on the example -- the example has to be well aligned to the LR image. In addition, the shallow model for patch blending made its performance highly dependent on how well the example could be aligned, limiting its adaptiveness to a more generic setting. 
	The proposed \alg~adopts the ideas of local texture matching and texture fusion like existing works, but we perform with high-level features and deep models targeting at the most generic scenario where the reference images can be arbitrary.   
	
	\section{Approach}
	\label{sec:approach}
	The reference-conditioned image super-resolution aims to estimate the SR image $I^{SR}$ from its LR counterpart $I^{LR}$ and the given reference image $I^{Ref}$, increasing plausible textures conditioned on the reference, while preserving the consistency to the LR image in color and content. 
	In the proposed \alg, beyond minimizing the distance between $I^{SR}$ and the original HR image $I^{HR}$ as most existing SR methods do, we further regularize on the texture consistency between $I^{SR}$ and $I^{Ref}$. The general objective could be expressed by
	\begin{equation}
	\hat{\theta} = \arg\underset{\theta}{\min}\frac{1}{n}\sum_{i=1}^{n}\left\{ \mathcal{L}_c\left( G_\theta(I_i^{LR}),I_i^{HR} \right) + \lambda\mathcal{L}_t\left( G_\theta(I_i^{LR}),I_i^{Ref} \right) \right\},
	\end{equation}
	where $G$ denotes the SR network with parameter $\theta$. $\mathcal{L}_c(\cdot, \cdot)$ and $\mathcal{L}_t(\cdot, \cdot)$ indicate the content loss and texture loss, respectively. For simplicity, assume each LR image corresponds to one reference, and there are $n$ LR images. 
	Section~\ref{subsec:losses} will discuss the loss functions in detail.
	\shrink
	\begin{figure}[ht]
		\centering
		\includegraphics[width=.8\columnwidth]{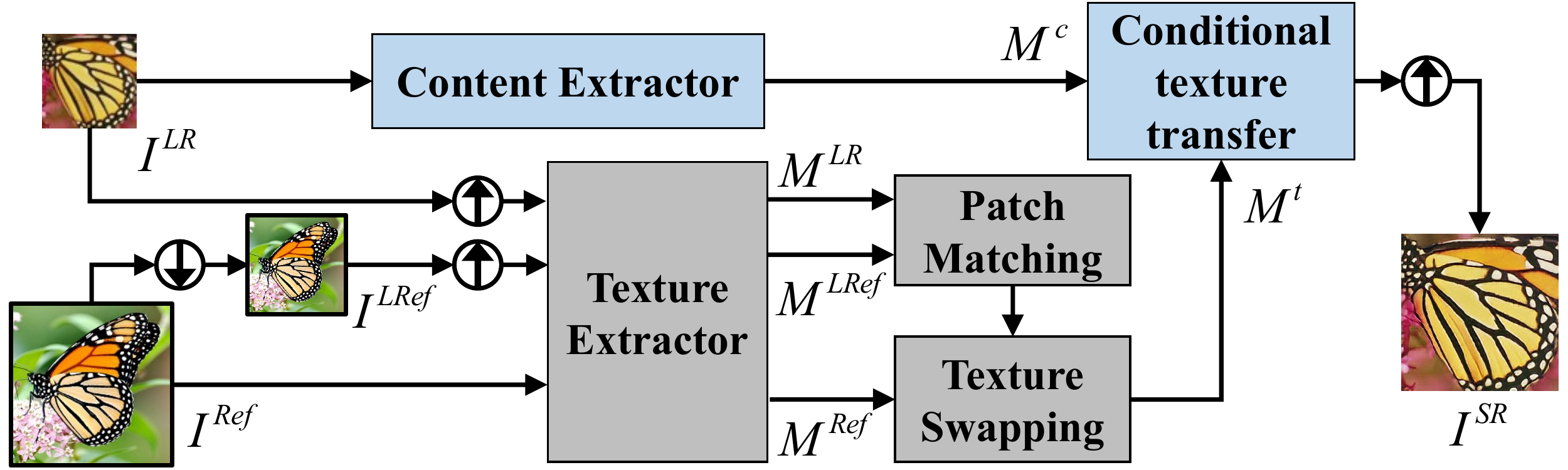}
		\caption{Overview of the proposed \alg. $I^{LR}$ and $I^{Ref}$ are the inputs to \alg. $I^{LR}$ is fed to the content extractor, obtaining the content feature map $M^c$. In parallel, $I^{Ref}$ and $I^{LR}$ are fed to the texture extractor after scale adjustment, yielding a series of feature maps for patch matching and texture swapping. The swapped texture map $M^t$ carries rich texture from $I^{Ref}$ while preserving the content structure of $I^{LR}$. Finally, $M^c$ and $M^t$ are adaptively fused through the conditional texture transfer, producing $I^{SR}$ with enhanced texture.
		}
		\label{fig:flow}
	\end{figure}
	\shrink
	
	An overall structure of the proposed \alg~is shown in Fig.~\ref{fig:flow}. 
	The main idea is to search for locally matched textures from the reference and adaptively transfer these textures to the SR image. 
	We design the structure as fusing two parallel streams, i.e., content and texture, which is consistent to the intuition of fusing texture to content. The content and texture are represented as high-level features extracted through deep models (i.e., content extractor and texture extractor, respectively), facilitating deep texture fusion which is more adaptive than using a shallow model. The content feature $M^c$ extracted from $I^{LR}$ and the texture feature $M^t$ extracted from $I^{Ref}$ are fused by the conditional texture transfer, which could learn to adaptively transfer perceptually consistent texture from $M^t$ to $M^c$. More details on conditional texture transfer will be discussed in Section~\ref{subsec:fusion}. 
	
	The content feature map $M^c$ is directly extracted from $I^{LR}$ through the content extractor. The corresponding texture feature map $M^t$ is obtained by local matching between the LR image $I^{LR}$ and reference $I^{Ref}$. Because $I^{LR}$ and $I^{Ref}$ may differ from each other in color and illumination, affecting texture matching and transfer, we choose to perform in the high-level feature space where most content structure as well as texture-related information are preserved. 
	Typically, the VGG19~\cite{simonyan2014very} model is adopted as the texture extractor, whose effectiveness on texture representation has been demonstrated by many empirical studies~\cite{gatys2015texture,gatys2016image}.
	To offset the bias from scale/resolution in patch matching, $I^{LR}$ and $I^{Ref}$ should be matched at similar scale. Intuitively, $I^{Ref}$ could be downsampled to the scale of $I^{LR}$, or $I^{LR}$ upsampled to the scale of $I^{Ref}$. The former is easier but the latter achieves more accurate matching with respect to location since matching is done at a larger scale. In addition, upscaling $I^{LR}$ preserves scale consistency with $M^{Ref}$, facilitating texture swapping from $M^{Ref}$ that carries richer texture information. 
	Therefore, we upscale $I^{LR}$ before feeding it to the texture extractor. However, we do not feed $I^{Ref}$ directly for patch matching because the upscaling process of $I^{LR}$ may introduce artifacts and blurry effects that would negatively affect the matching result. Hence, we downscale $I^{Ref}$ and followed by the same upscaling process as that of $I^{LR}$ to achieve more accurate patch matching. Typically, the downscaling uses bicubic interpolation, while the upscaling could be an existing SR method or even bicubic interpolation.  
	Section~\ref{subsec:swap} will further detail the patching matching and texture swapping. 
	
	
	\subsection{Patch matching and texture swapping}  
	\label{subsec:swap}
\begin{figure}[ht]
		\centering
		\includegraphics[width=.6\columnwidth]{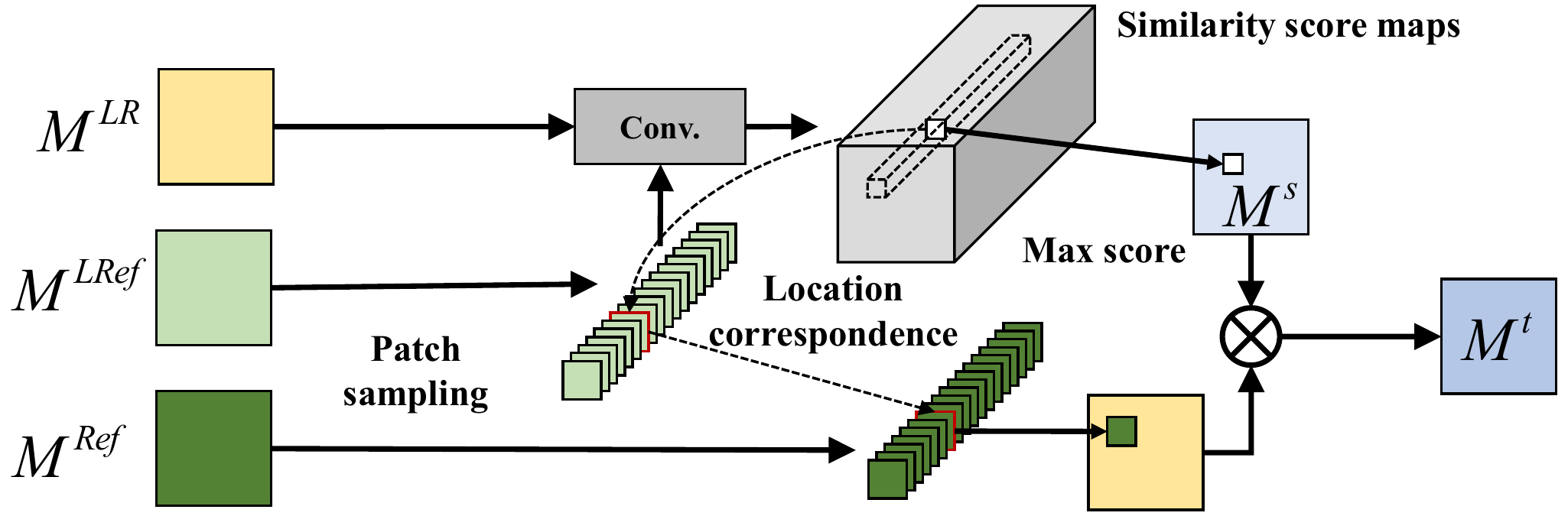}
		\caption{Patch matching and texture swapping.}
		\label{fig:flow_2}
	\end{figure}

	Patch matching and texture swapping generate the texture map $M^t$ that carries rich texture information from $I^{Ref}$ while preserving the content structure of $I^{LR}$. The details are shown in Fig.~\ref{fig:flow_2}. For simplicity, all feature maps are shown as single channel. 
	\shrink
	\shrink
	
	Patch matching is first performed between $M^{LR}$ and $M^{LRef}$. 
	The patch-wise similarity is measured by inner product, 
	\begin{equation}
	s_{i,j} = \left< \frac{p^{LR}_i}{\|p^{LR}_i\|}, \frac{p^{LRef}_j}{\|p^{LRef}_j\|} \right> 
	\label{eq:inner}
	\end{equation} 
	where $s_{i,j}$ denotes the similarity score between the $i$th patch from $M^{LR}$ (i.e., $p_i^{LR}$) and the $j$th patch from $M^{LRef}$ (i.e., $p_j^{LRef}$). 
	As shown in Fig.~\ref{fig:flow_2}, $M^{LRef}$ is reshaped to a sequence of patches by dense sampling. The patches can be considered as kernels, hence the inner product can be approximated by performing convolution  between $M^{LR}$ and the patch kernels, to yield a sequence of similarity score maps. The maximum score at each pixel location across the similarity score maps indicates the best matched $M^{LRef}$ patch.
	The similarity map $M^s$ records those maximum scores with structural identity to $M^{LR}$. Since $M^{Ref}$ is structurally identical to $M^{LRef}$, the patch matching correspondence between $M^{LR}$ and $M^{LRef}$ is identical to that between $M^{LR}$ and $M^{Ref}$. Therefore, texture-richer patches from $M^{Ref}$ is swapped to $M^{LR}$ according to the correspondence. The overlaps between swapped patches are averaged.
	Considering the uncorrelated texture that may degrade the SR performance, the swapped texture map is multiplied by $M^s$, which weights down those uncorrelated texture because of its low similarity score. This significantly boost the adaptiveness of texture transfer, which will be demonstrated in Section~\ref{subsec:weight_effect}. Finally, we obtain a weighted texture map $M^t$. 

	\subsection{Conditional texture transfer}  
	\label{subsec:fusion}
\begin{figure}[ht]
		\centering
		\includegraphics[width=.7\columnwidth]{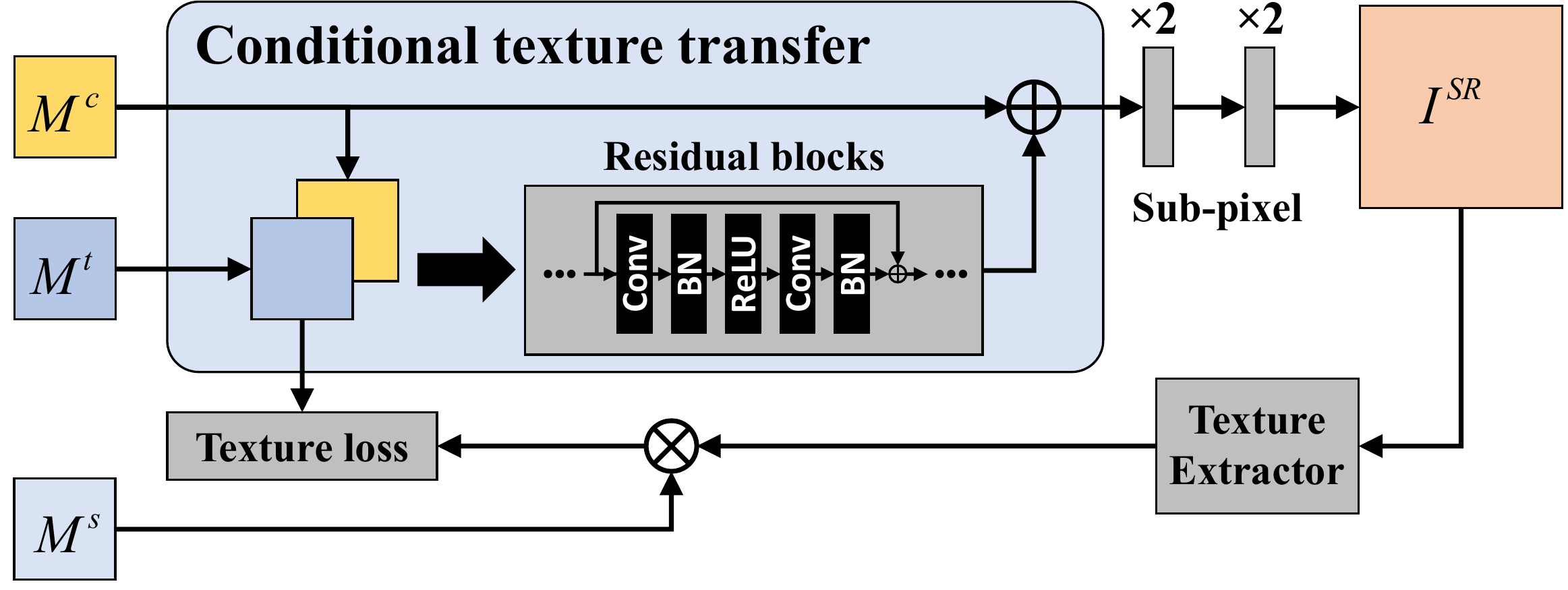}
		\caption{The pipeline of conditional texture transfer. The transfered feature map is fed to the sub-pixel layers for upscaling. The texture loss is is computed between $M^t$ and the weighted feature map extracted from $I^{SR}$.}
		\label{fig:flow_3}
	\end{figure}

	Based on the content feature map $M^c$ and weighted texture map $M^t$, the conditional texture transfer would adaptively fuse textures from $M^t$ to $M^c$ by a deep residual network. 
	The pipeline of conditional texture transfer is shown in Fig.~\ref{fig:flow_3},
	\shrink
	\shrink
	where $\oplus$ and $\otimes$ represent element-wise summation and multiplication, respectively. Since the texture transferred to $I^{SR}$ is supposed to be visually consistent with $I^{LR}$, it is necessary to transfer $M^t$ conditioned on $M^c$. As shown in the blue block of Fig.~\ref{fig:flow_3}, $M^c$ is concatenated to $M^t$ as the condition, and they are fed to the deep residual blocks~\cite{he2016deep}, which learn to adaptively extract consistent texture from $M^t$ conditioned on $M^c$. The extracted texture is then added to $M^c$.
	Finally, sub-pixel convolution~\cite{shi2016real} is employed as the upscaling process, which is beneficial in both accuracy and speed.
	
	Different from traditional SISR methods that only focus on losses between $I^{SR}$ and the ground truth, the reference-conditioned SR also takes into account the loss between $I^{SR}$ and $I^{Ref}$, which we refer to as the texture loss. 
	Specifically, the texture loss is calculated between $M^{t}$ derived from $I^{Ref}$ and the weighted feature map extracted from $I^{SR}$ with $M^s$ being the weight. 
	More details on loss functions will be discussed in Section~\ref{subsec:losses}. 
	\subsection{Content loss $\boldsymbol{\mathcal{L}_c}$ and texture loss $\boldsymbol{\mathcal{L}_t}$}  
	\label{subsec:losses}
	As briefly discussed at the beginning of Section~\ref{sec:approach}, in order to preserve the structural information of the LR image, improve the visual quality of the SR image, as well as taking advantage of the rich texture details from the reference image, the objective function we develop involves both content loss $\mathcal{L}_c$ and texture loss $\mathcal{L}_t$. The content loss is three-fold, including the reconstruction loss, $\mathcal{L}_{rec}$, to preserve the structural information, the perceptual loss, $\mathcal{L}_{per}$, and the adversarial loss, $\mathcal{L}_{adv}$, to boost the visual quality. The texture loss is added for the network to adaptively enhance the texture transferred from the reference image. 
	
	\textbf{Reconstruction loss} is widely adopted in most SR works. To achieve the objective of obtaining higher PSNR, MSE is usually used to measure the reconstruction loss. In this paper, we adopt the $\ell_1$-norm,
	\begin{equation}
	\mathcal{L}_{rec} = \frac{1}{HW}\sum_{x=1}^{H}\sum_{y=1}^{W}|I^{HR}_{x,y}-I^{SR}_{x,y}|,
	\label{eq:l_rec}
	\end{equation} 
	where $H$ and $W$ denote the height and width of the HR/SR image, respectively. The $\ell_1$-norm would further sharpen $I^{SR}$ as compared to MSE. In addition, it is consistent to the objective of WGAN-GP, which will be discussed later in the adversarial loss.
	
	\textbf{Perceptual loss} has been investigated in recent SR works~\cite{bruna2016super,johnson2016perceptual,ledig2017photo,sajjadi2017enhancenet} for better visual quality. We calculate the perceptual loss based on the relu5\_1 layer of VGG19~\cite{simonyan2014very}, 
	\begin{equation}
	\mathcal{L}_{per}=\frac{1}{V}\sum_{i=1}^{C}\left\|\phi_i(I^{HR})-\phi_i(I^{SR})\right\|_F,
	\end{equation}
	where $V$ and $C$ indicate the tensor volume and channel number of the feature maps, respectively, and $\phi_i$ denotes the $i$th channel of the feature maps extracted from the hidden layer of VGG19 model. $\|\cdot\|_F$ denotes the Frobenius norm.   
	
	\textbf{Adversarial loss} could significantly enhance the sharpness/visual quality of $I^{SR}$. 
	Here, we adopt WGAN-GP~\cite{gulrajani2017improved}, which improves upon WGAN by penalizing the gradient, achieving more stable results. Because the Wasserstein distance in WGAN is based on $\ell_1$-norm, we also use $\ell_1$-norm as the reconstruction loss (Eq.~\ref{eq:l_rec}). 
	Intuitively, consistent objectives in optimization would help yield more stable results. 
	The adversarial loss and objective of WGAN are expressed as
	\begin{equation}
	\mathcal{L}_{adv} = -\mathbb{E}_{\tilde{x}\sim\mathbb{P}_g}[D(\tilde{x})],\;\;\;
	\underset{G}{\min}\;\underset{D\in\mathcal{D}}{\max}\;\mathbb{E}_{x\sim\mathbb{P}_r}[D(x)]-\mathbb{E}_{\tilde{x}\sim\mathbb{P}_g}[D(\tilde{x})], 
	\end{equation}
	where $\mathcal{D}$ is the set of 1-Lipschitz functions, and $\mathbb{P}_r$ and $\mathbb{P}_g$ are the model distribution and real data distribution, respectively.   
	
	\textbf{Texture loss} is built on the idea of Gatys et al.~\cite{gatys2015texture,gatys2016image}, where Gram matrix was applied to statistically preserve the texture from the style image. 
	To preserve the consistency between the content and transferred texture, the similarity map $M^{s}$ is utilized, as illustrated in Fig.~\ref{fig:flow_3}, to suppress uncorrelated texture. The texture loss is written as
	\begin{equation}
	\mathcal{L}_t = \frac{1}{4V^2}\left\| Gr(\phi(I^{SR})\otimes M^s) - Gr(M^t) \right\|_F,
	\end{equation} 
	where $Gr(\cdot)$ denotes the operator that computes the Gram matrix, and $\phi(\cdot)$ indicates the feature maps from the relu3\_1 layer of VGG19 model, whose scale is the same as that of the texture feature map $M^{t}$. $V$ is the volume of the feature map tensor from VGG19, and $\otimes$ denotes the element-wise multiplication.
	\section{Dataset}
	\label{sec:dataset}
	For reference-based SR problems, the similarity between the LR image and reference affects SR results significantly. In general, references with various similarity levels to the corresponding LR image should be provided for the purpose of learning and evaluating the adaptiveness to similar and dissimilar textures in references. We refer to pairs of LR and reference images as LR-Ref pairs. 
	
	
	We use SIFT~\cite{lowe1999object} features to measure the similarity between two images because SIFT features characterize local texture information that is in line with the objective of local texture matching. In addition, SIFT feature matching is conducted 
	in the pixel space which is more rigorous than high-level feature matching, providing more visually correlated pairs. 
	
	We build the training and validation datasets based on CUFED~\cite{Wang_16_CVPR} that contains 1,883 albums. We choose to use album images since images collected from the same event are supposed to be taken in similar environment. Each album describes one of the 23 most common events in our daily life, ranging from Wedding to Nature Trip. The size of albums varies between 30 and 100 images. Within each album, we collect image pairs in different similarity levels based on SIFT feature matching. We quantitatively collect five similarity levels denoted as XH (extra-high), H (high), M (medium), L (low), and XL (extra-low). From each paired images, we randomly crop 160$\times$160 patches from one image as the original HR images (LR images are obtained by downscaling), and the corresponding references are cropped from the other image. 
	In this way, we collect 13,761 LR-Ref pairs for training purpose. It is worth noting that in  
	building the training dataset, it is not necessary to present references at all five similarity levels for `each' LR image although the training set as a whole should contain LR-Ref pairs at different similarity levels. 
	
\begin{figure}[ht]
		\centering
		\begin{tabular}{C{2.2cm}C{1.65cm}C{1.65cm}C{1.65cm}C{1.65cm}C{1.65cm}C{.2cm}}
			HR & XH & H & M & L & XL & ~
		\end{tabular}
		
		\includegraphics[width=.9\columnwidth]{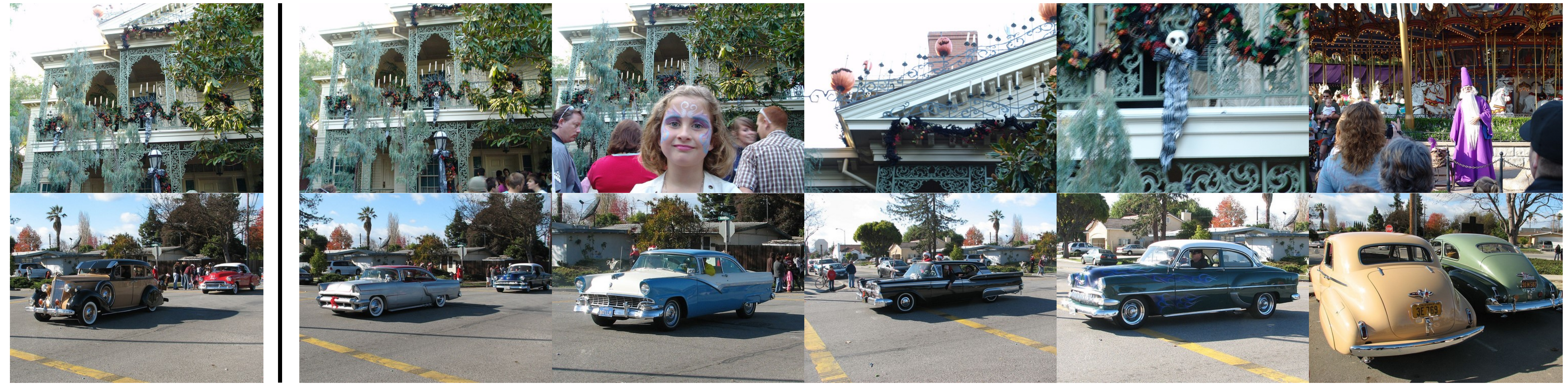}
		\caption{Examples from the \dataset~dataset. The left column is the HR image for testing. The right columns are corresponding references in five similarity levels, i.e., extra-high (XH), high (H), medium (M), low (L), and extra-low (XL).}
		\label{fig:test_set}
	\end{figure}

	On the other hand, the validation dataset does need each LR image to have references at all five similarity levels in order to extensively evaluate the adaptiveness of the network to references with different similarity levels. 
	We use the same way to collect image pairs as in building the training dataset. In total, the validation set contains 126 groups of testing samples with each group consisting of one HR image and five references at five different similarity levels. Some randomly selected testing groups are shown in Fig.~\ref{fig:test_set}. 
	We refer to the collected training and validation sets as \dataset. The construction of \dataset~ largely facilitates performance evaluation of the proposed reference-conditioned SR research, providing a benchmark for the study of reference-based SR in general.
	\shrink
	\shrink
	
	\shrink
	\section{Experimental Results}
	\label{sec:experiment}
	In this section, both quantitative and qualitative comparisons are conducted to demonstrate the effectiveness of the proposed \alg~in boosting SR performance in aspects of visual quality, texture richness, and content consistency. 
	
	\subsection{Training details and parameters}
	In training, the LR images are obtained by downscaling (4$\times$) the HR images through bicubic interpolation, thus the LR images are of the size $40\times 40$. The corresponding reference is fed with the original size, 160$\times$160. All feature maps keep the same size of 40$\times$40, to facilitate patch matching, texture swapping, and concatenation of content and texture maps. The weights parameters for $\mathcal{L}_{per}$, $\mathcal{L}_{adv}$, and $\mathcal{L}_t$ are $\alpha$=1e-4, $\beta$=1e-6, and $\lambda$=1e-4, respectively. Adam optimizer is used with the initial learning rate of 1e-4. The network is pre-trained for 5 epochs, where only $\mathcal{L}_{rec}$ is applied. Then, all losses are involved to train another 100 epochs, during which the learning rate is decayed by 0.1 for each 50 epochs. Note that $\mathcal{L}_{t}$ is only applied on the conditional texture transfer network. The whole framework could be trained end-to-end. However, the patch matching is time-consuming, occupying over 90\% run time during training. Hence we calculate $M^t$ offline for each training pair because the process of generating $M^{t}$ only involves the pre-trained VGG19 model. To further speed up the training process, we could use a pre-trained SR network, e.g., MSE-based EDSR or GAN-based SRGAN, as the content extractor to obtain $M^c$. 
	
	\subsection{Quantitative evaluation}
	We compare the PSNR and SSIM with other SR methods, as shown in Table~\ref{tab:cmp}. SelfEx~\cite{Huang-CVPR-2015} is a non-learning-based method using the LR input itself as reference. SRCNN~\cite{dong2014learning}, SCN~\cite{wang2015deep}, DRCN~\cite{kim2016deeply}, LapSRN~\cite{zhu2014fast}, and EDSR~\cite{lim2017enhanced} are learning-based methods by minimizing MSE. ENet~\cite{sajjadi2017enhancenet} and SRGAN~\cite{ledig2017photo} are also learning-based but further utilize the perceptual-related constraints to enhance the visual quality. Landmark~\cite{yue2013landmark} is a reference-based SR method, retrieving references that could be well-aligned to $I^{LR}$. The ``\alg-'' denotes a simplified version of \alg~ by removing the adversarial loss, which is supposed to achieve comparable PSNR as the MSE-based methods. All methods are tested on the \dataset~dataset. The Landmark method is tested with the reference of similarity level M. Our methods are tested with each of the five references and results averaged. This individual performance is listed in Table~\ref{tab:cmp2}. 
	\shrink
	\begin{table}[ht]
		\centering
		\caption{Comparison of different SR methods in PSNR and SSIM.}
		\label{tab:cmp}
		\begin{tabular}{C{1.cm}|C{1.7cm}C{1.7cm}C{1.7cm}C{1.7cm}C{1.7cm}C{1.7cm}}
			\hline
			~ & Bicubic  & SRCNN & SCN & DRCN & LapSRN & EDSR  \\
			\hline
			PSNR & 24.18 & 25.33 & 25.45 & 25.26 & 24.92 & \textbf{26.81}  \\
			SSIM & .6837 & .7451 & .7426 & .7337 & .7299 & \textbf{.7923}  \\
			\hline
			\hline
			~  & SelfEx & Landmark & ENet & SRGAN & \alg - & \alg \\
			\hline
			PSNR & 23.22 & 24.91 & 24.24 & 24.40 & 26.23 & 24.60  \\
			SSIM & .6799 & .7176 & .6948 & .7021 & .7737 & .7086 \\
			\hline
		\end{tabular}
		\caption{PSNR and SSIM of \alg~with different reference levels.}
		\label{tab:cmp2}
		\begin{tabular}{C{1.cm}|C{1.7cm}C{1.7cm}C{1.7cm}C{1.7cm}C{1.7cm}C{1.7cm}}
			\hline
			~  & XH & H & M & L & XL & Average \\
			\hline
			PSNR & 24.57 & 24.58 & 24.56 & 24.63 & \textbf{24.69} & 24.60  \\
			SSIM & .7099 & .7082 & .7075 & .7079 & \textbf{.7094} & .7086 \\
			\hline
		\end{tabular}
		\caption{Texture distance to the reference based on Gram matrix.}
		\label{tab:cmp3}
		\begin{tabular}{C{1.cm}|C{1.7cm}C{1.7cm}C{1.7cm}C{1.7cm}C{1.7cm}C{1.7cm}}
			\hline
			~ & Landmark  & EDSR & ENet & SRGAN & \alg- & \alg \\
			\hline
			Gram & 55.25 & 37.00 & 27.26 & 32.21 & 33.72 & \textbf{22.77}  \\
			\hline
		\end{tabular}
		
	\end{table}
	\shrink
	
	From PSNR and SSIM, we observe that those methods minimizing MSE perform better than GAN-based methods. However, PSNR and SSIM cannot adequately reflect visual quality of the image, especially when the upscaling factor is relatively large (e.g., 4$\times$) and/or the image is originally with rich texture. The GAN-based methods would present more details that may deviate from the content of the original image, or even introduce artifacts to make the generated images look sharper. 
	Note that \alg- achieves comparable PSNR and SSIM with the state-of-the-art, i.e., EDSR. In the individual performance of \alg~with each reference level, 
	the highest PSNR/SSIM is achieved at XL because the texture from reference is mostly suppressed, yielding relatively smooth results. By contrast, the score of XH is lower because correlated texture are transferred to the SR results which may deviate from the original HR image. 
	
\begin{figure}[ht]
		\centering
		\begin{tabular}{C{2.6cm}C{2.6cm}C{2.6cm}C{2.6cm}}
			Landmark & EDSR & \alg- & Original \\
			\hline
			ENet & SRGAN & \alg & Reference\\
			\hline
		\end{tabular}
		\includegraphics[width=.9\columnwidth]{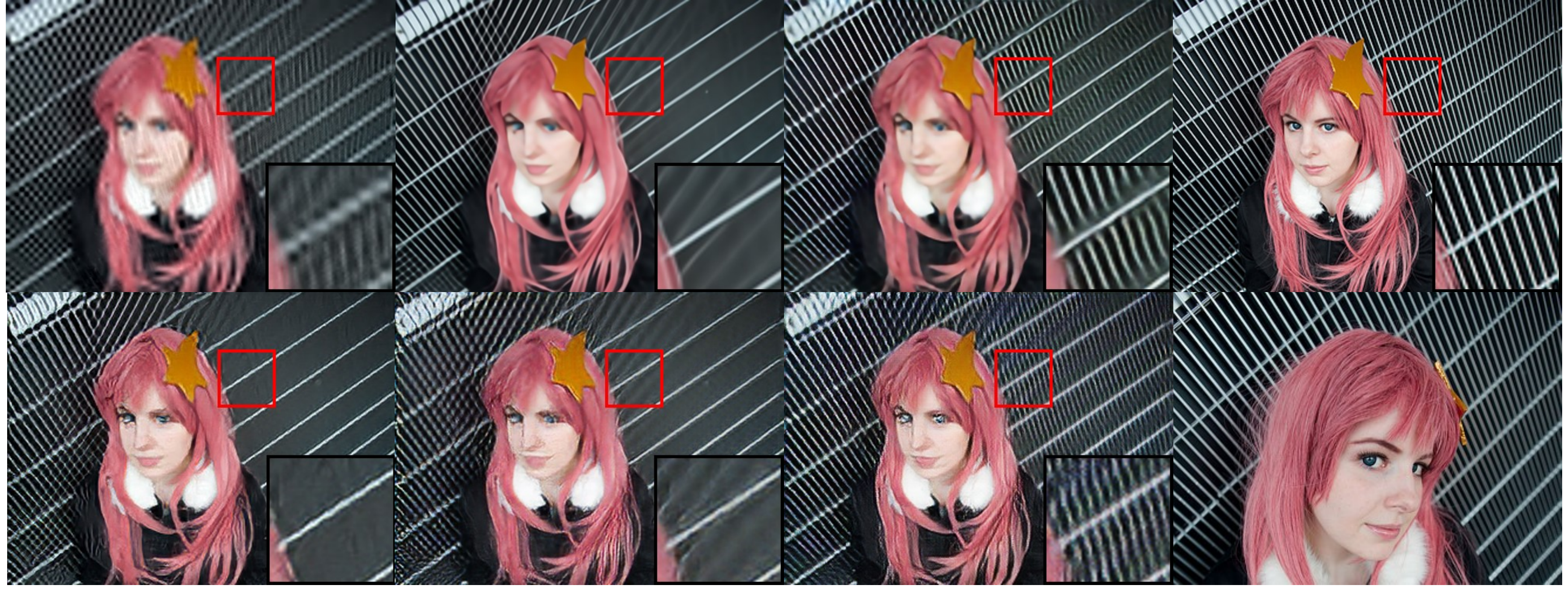}
		\includegraphics[width=.9\columnwidth]{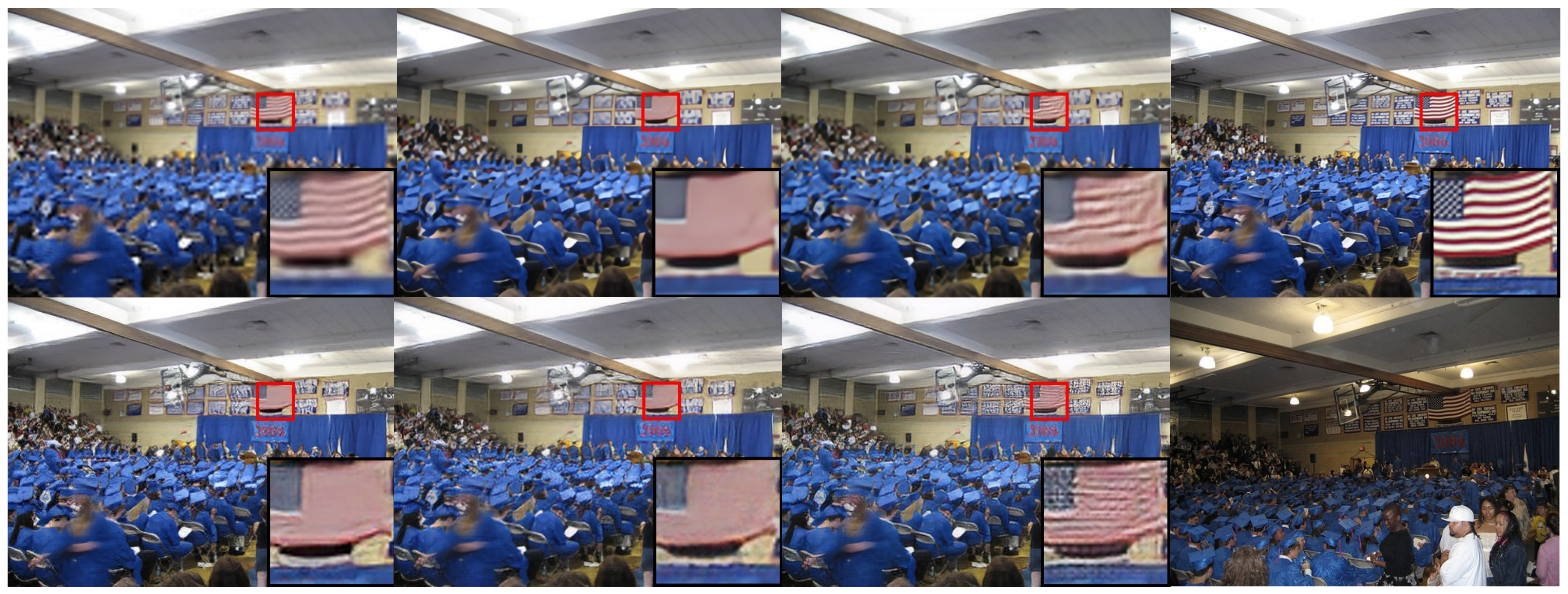}
		\includegraphics[width=.9\columnwidth]{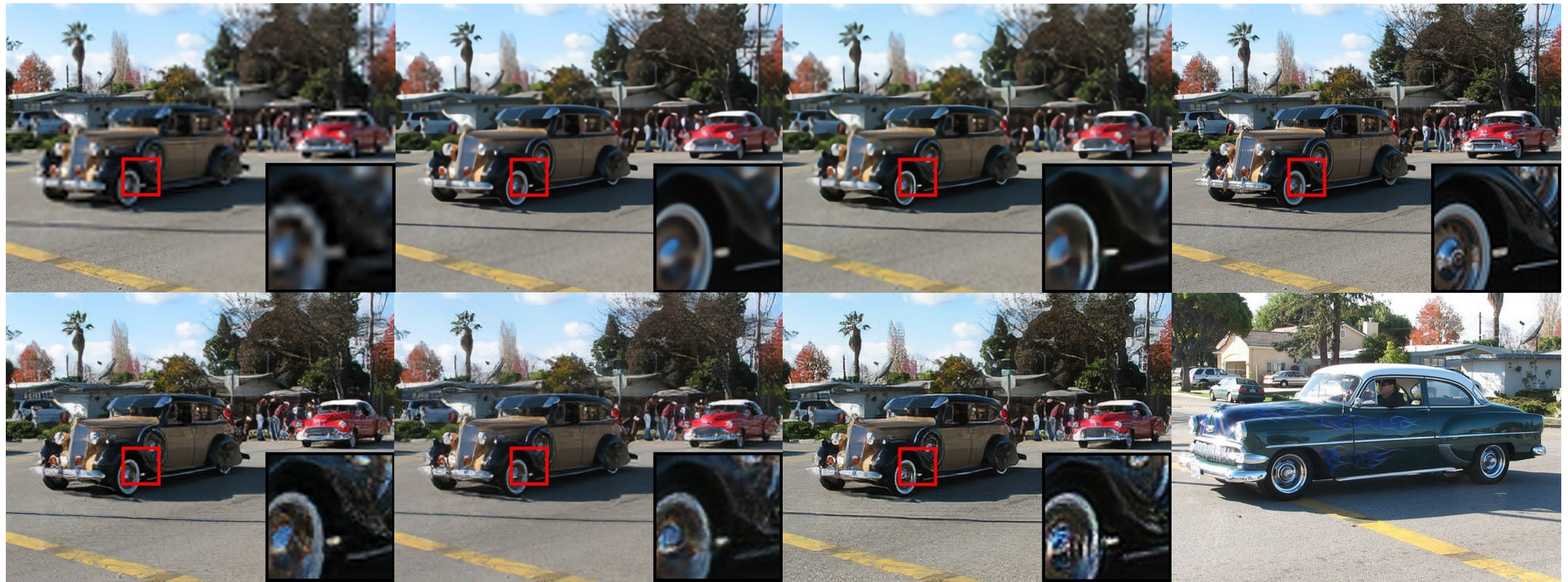}
		\caption{Visual comparison to other SR methods.}
		\label{fig:cmp}
	\end{figure}

	To illustrate the conditioned texture transfer, we measure the textural similarity between $I^{SR}$ and $I^{Ref}$ (level M) based on the Gram matrix as shown in Table~\ref{tab:cmp3}. \alg~presents the smallest distance since it transfers texture from the reference. ENet also achieves a relatively small distance because it also regularizes on texture but between $I^{SR}$ and the original HR image.   
	Some typical testing results are shown in Fig.~\ref{fig:cmp},
	where \alg~and \alg- present more textures that are transferred from the references. We observe that Landmark could well utilize the texture from references only when the reference can be well aligned to the input. For example, Landmark can better recover the flag since it can find well aligned patch in the reference. For the other examples, Landmark fails to recover details. Our method could better tolerate misaligned or unrelated texture. The MSE-based methods tend to generate clean but blurry images. By contrast, \alg- gains rich textures from references. The GAN-based methods present better visual quality, but still cannot recover plausible textures like \alg.
	\shrink
	\shrink
	\begin{figure}[ht]
		\centering
		\includegraphics[width=.85\columnwidth, trim=60 5 80 0, clip]{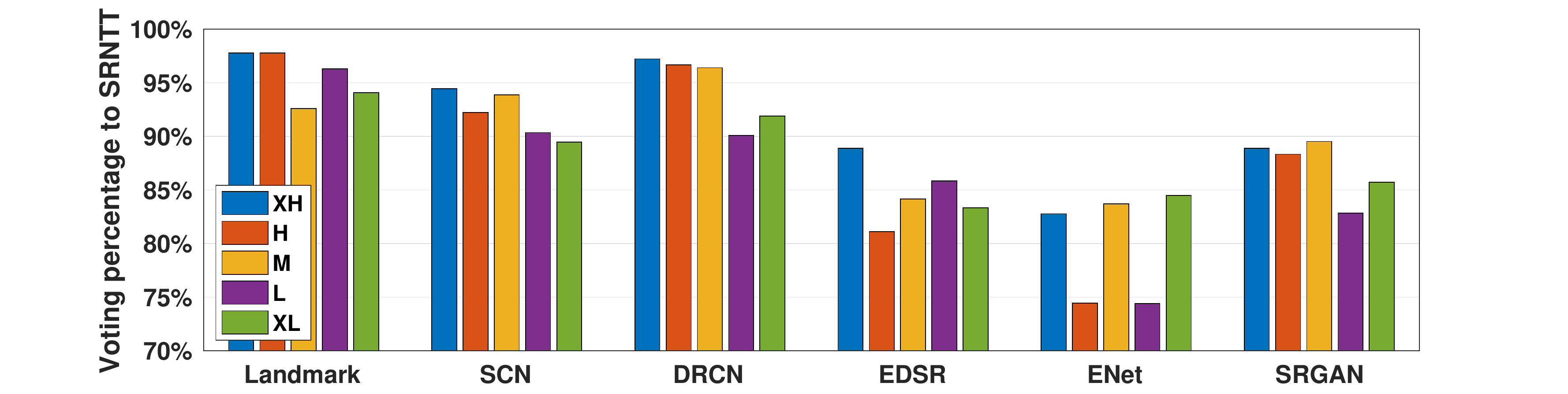}
		\caption{Percentage of votes of \alg~as compared to each of other methods. }
		\label{fig:votes_bar}
	\end{figure}
	\begin{figure}[ht]
		\centering
		\includegraphics[width=.85\columnwidth]{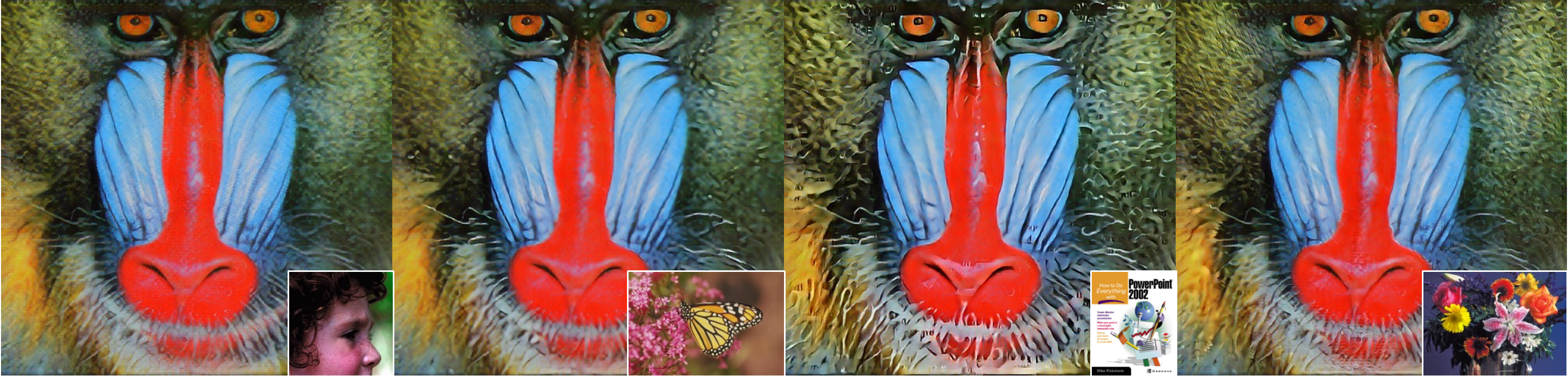}
		\caption{The texture is enhanced conditioned on the reference that is shown in the bottom-right corner. Please zoom in for better view.}
		\label{fig:conditions}
		\shrink
	\end{figure}
	\shrink
	
	\subsection{Qualitative evaluation by user study}
	To evaluate the visual quality of the SR images, we conduct user study which is widely adopted in many GAN-related works~\cite{ledig2017photo,sajjadi2017enhancenet}. \alg~ is compared to other six methods, i.e., Landmark, SCN, DRCN, EDSR, ENet, and SRGAN. 
	We present the users with pair-wise comparisons, i.e., \alg~vs. other, and ask the users to select the one with better visual quality and more natural looking. For each reference level, 1,890 votes are collected on the testing results randomly selected from the \dataset~dataset. 
	Fig.~\ref{fig:votes_bar} shows the voting results,  
	where the percentages of votes of \alg~ as compared to other methods, and demonstrate a roughly descending trend as the references become less similar to the LR image. 
	
	%
	
		\subsection{Texture transfer results}
	More general but still extreme for existing reference-based SR methods, the reference could be an arbitrary image, which may significantly deviate from $I^{LR}$ in aspect of content and texture. 
	An example is shown in Fig.~\ref{fig:conditions}, where the SR results are conditioned on the references shown in the bottom-right corner.
	\alg~could adaptively transfer correlated textures to $I^{SR}$, thus presenting enhanced texture conditioned on the reference. Comparing the first and last results, the latter is visually sharper because its reference carries richer texture and with higher resolution. The third result shows strong edge as the reference.

	
	\subsection{Investigation on extreme conditions} 
	\label{subsec:weight_effect} 
	\begin{figure}[ht]
		\centering
		\begin{tabular}{C{1.6cm}C{2cm}C{1.5cm}C{1.8cm}C{2.6cm}}
			Dark & Bright & Noise & Original & ~
		\end{tabular}
		\subfloat[Our results with extreme references]
		{
			\rotatebox{90}{\begin{tabular}{C{1.6cm}C{2cm}C{1.5cm}}
					\alg & \alg$/M^s$ & References
			\end{tabular}}
			\includegraphics[width=.6\columnwidth]{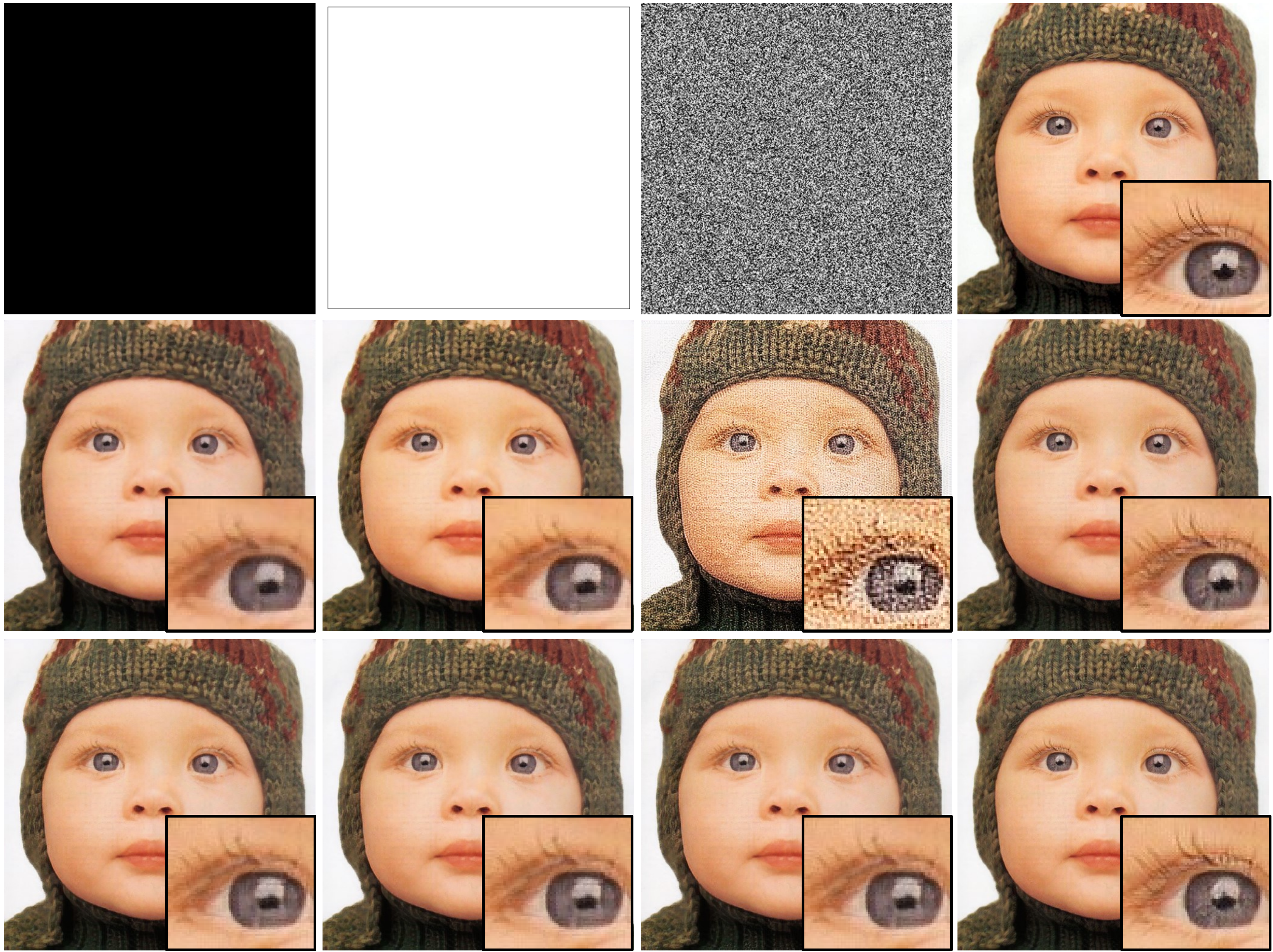}
			\label{subfig:weight1}
		}
		\hspace{.5cm}
		\subfloat[Baselines]
		{
			\rotatebox{90}{\begin{tabular}{C{1.6cm}C{2cm}C{1.5cm}}
					SRGAN & ENet & EDSR 
			\end{tabular}}
			\includegraphics[width=.15\columnwidth]{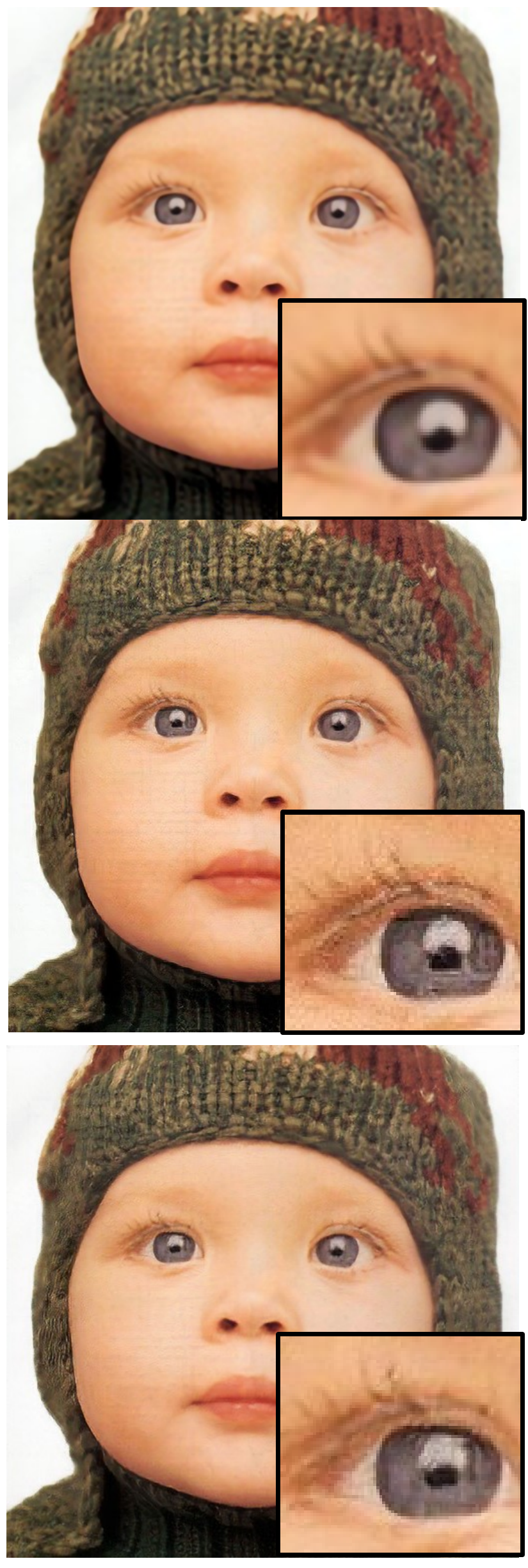}
			\label{subfig:weight2}
		}
		\caption{(a) The extreme references and the corresponding SR results from the proposed \alg~and \alg/$M^s$ which is a trimmed version of \alg~by removing $M^s$. (b) Baseline results for comparison purpose.}
		\label{fig:weight}
	\end{figure}
	This section investigates extreme cases in reference-conditioned SR where the reference is simply image with homogeneous intensity levels or even random noise.
	The proposed \alg~can cope with these extreme cases by introducing the similarity map $M^s$, which could effectively suppress unrelated textures. We also train \alg~without $M^s$, which is referred to as \alg/$M^s$. Without considering the extreme references, \alg~and \alg/$M^s$ show similar performance. However, when given extreme references, \alg/$M^s$ may introduce negative effects from the reference as shown in Fig.~\ref{fig:weight}.
	Given all-dark or all-bright references, which do not provide any extra texture information to the conditional texture transfer network, the results of \alg~and \alg$/M^s$ are close to the state-of-the-art GAN-based methods, i.e., ENet and SRGAN. However, given the random noise as reference, \alg$/M^s$ transfers the texture of noise to the SR image by mistake, while such uncorrelated texture is successfully suppressed by \alg, demonstrating the adaptiveness gained from $M^s$. When the reference is perfect, i.e., the original image, the results from both \alg~ and \alg$/M^s$ show much finer details. Therefore, $M^s$ plays a critical role in suppressing uncorrelated textures while encouraging correlated ones.
	\shrink

	\section{Conclusion}
	\label{sec:conclusion}
	This paper exploited the reference-conditioned solution for solving SR problems where the reference can be an arbitrary image. 
	We proposed \alg, an end-to-end network structure that performs adaptive texture transfer from the reference to recover more plausible texture in the SR image. Both quantitative and qualitative experiments were conducted to demonstrate the effectiveness and adaptiveness of \alg, even with extreme cases of references. In addition, a new dataset \dataset~was constructed to facilitate the evaluation of reference-conditioned SR methods. It also provides a benchmark for future reference-based SR research in general.     
	
\clearpage

\bibliographystyle{splncs03}
\bibliography{egbib}

\end{document}